\documentclass[fleqn,10pt]{wlscirep}
\usepackage[utf8]{inputenc}
\usepackage[T1]{fontenc}
\usepackage{enumitem}

\title{SME-TEAM: Leveraging Trust and Ethics for Secure and Responsible Use of AI and LLMs in SMEs}

\author[1,*]{Iqbal H. Sarker}
\author[1]{Helge Janicke}
\author[1]{Ahmad Mohsin}
\author[2]{Leandros Maglaras}
\affil[1]{Centre for Securing Digital Futures, Edith Cowan University, Perth, WA-6027, Australia.}
\affil[2]{School of Computing, De Montfort University, UK.}

\affil[*]{Correspondence: m.sarker@ecu.edu.au}

\begin{abstract}
	Artificial Intelligence (AI) and Large Language Models (LLMs) are revolutionizing today's business practices; however, their adoption within small and medium-sized enterprises (SMEs) raises serious trust, ethical, and technical issues. In this perspective paper, we introduce a structured, multi-phased framework, ``SME-TEAM" for the secure and responsible use of these technologies in SMEs. Based on a conceptual structure of four key pillars, i.e., Data, Algorithms, Human Oversight, and Model Architecture, SME-TEAM bridges theoretical ethical principles with operational practice, enhancing AI capabilities across a wide range of applications in SMEs. Ultimately, this paper provides a structured roadmap for the adoption of these emerging technologies, positioning trust and ethics as a driving force for resilience, competitiveness, and sustainable innovation within the area of business analytics and SMEs. \\

\end{abstract}

\begin{document}

\flushbottom
\maketitle
%

\section{Introduction}
\begin{figure}[ht]
	\centering
	\includegraphics[width=.8\linewidth, keepaspectratio]{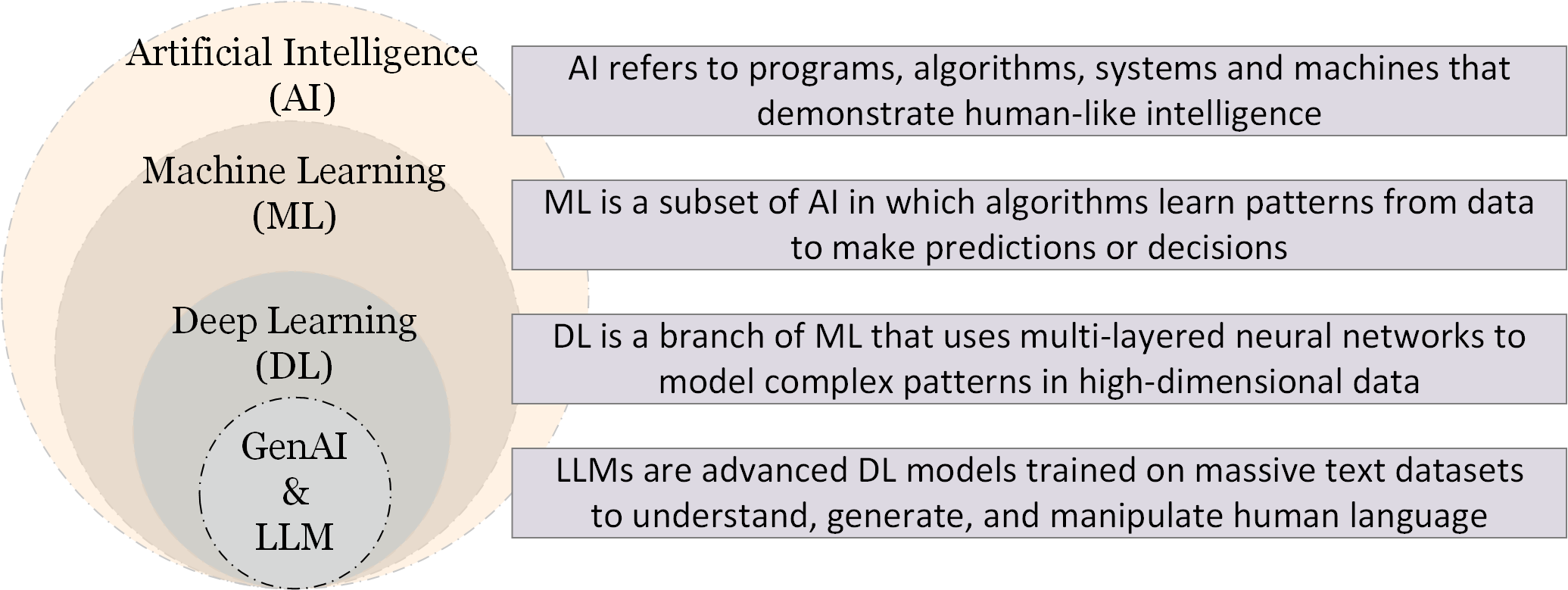}
	\caption{Typically involved technologies such as machine learning, deep learning and large language models in the broad area of AI.}
	\label{fig:AI-LLM-position}
\end{figure}

SMEs are the backbone of national and global economies, representing over 90\% of businesses worldwide, around 50\% of global employment, and thus playing a significant role in economic sustainability as well as employment generation \cite{chidukwani2022survey, soomro2025sem}. In Australia, SMEs represent over 98\% of all businesses and contribute nearly one-third of GDP \cite{SMEAustralia2023}. Similarly, SMEs dominate the European economy \cite{jafarzadeh2024supporting}. With the rapid advancement of AI technologies like machine learning (ML), deep learning (DL), and large language models (LLMs) \cite{sarker2024ai, wei2022artificial, oldemeyer2025investigation}, as defined and shown their relationship in Figure \ref{fig:AI-LLM-position} - SMEs are undergoing digital transformation. These emerging technologies enable innovative business models, enhance productivity and customer engagement, intelligent decision-making as well as market expansion with new employment opportunities. Nowadays, digital business transformation with the integration of these emerging technologies is reshaping SMEs across diverse sectors - such as manufacturing \cite{chen2024digital, qureshi2025analysing, graziosi2024vision}, healthcare \cite{treasure2019scaffolded}, agriculture and food \cite{el2024leveraging, gupta2025unveiling}, finance \cite{vukovic2025ai, lu2023developing}, and retail \cite{razzaq2025empowering}. By enabling business capabilities, this transformation positions SMEs as pivotal contributors to growth and resilience in the rapidly evolving digital economy worldwide.

While AI technologies with digital transformation create immense opportunities in today's business, SMEs face several challenges that complicate its adoption \cite{khanal2024building}. For instance, technical limitations include the lack of explainability in AI models and vulnerability to errors or adversarial attacks, which reduces trust in model outcomes and decision-making. Ethical risks include bias, concerns about fairness, and a lack of transparency, which threaten customer confidence and brand reputation. Implementation costs and shortages of AI-skilled professionals constrain effective deployment and oversight. From the security point of view, digital transformation and AI systems also introduce risks of cyberattacks, data breaches, and adversarial manipulation that may threaten business continuity \cite{ode2025social}. Moreover, weak governance structures, limited regulatory awareness, and a lack of clear mechanisms for accountability impede the embedding of responsible AI practices in SMEs. These risks collectively endanger customer trust and compliance, exposing SMEs to reputational damage, financial losses, and reduced competitiveness in the evolving digital economy.

\begin{figure}[ht]
	\centering
	\includegraphics[width=.6\linewidth, keepaspectratio]{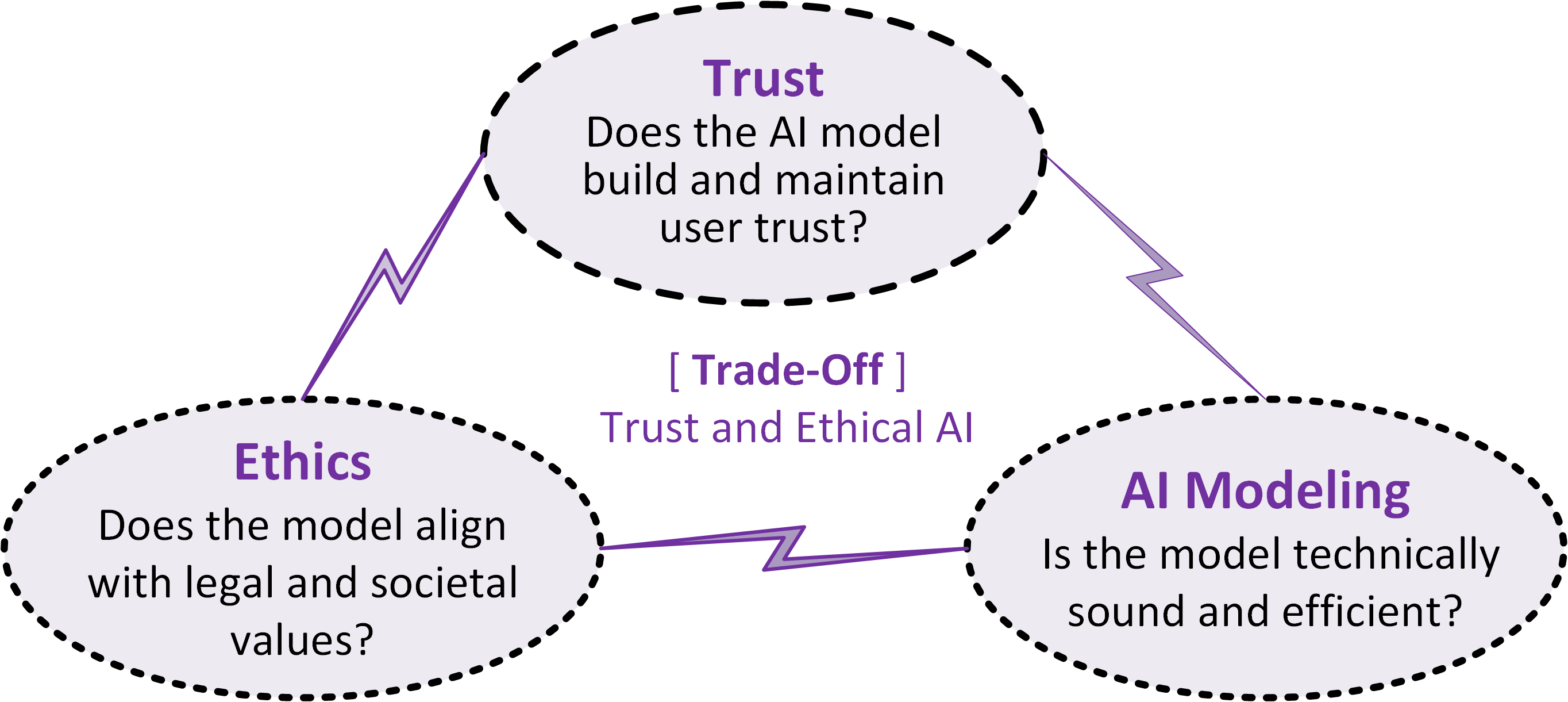}
	\caption{An illustration of the key aspects - Trust, Ethics and AI Modeling in SME-TEAM framework.}
	\label{fig:trade-off}
\end{figure} 

\begin{figure}[ht]
	\centering
	\includegraphics[width=.8\linewidth, keepaspectratio]{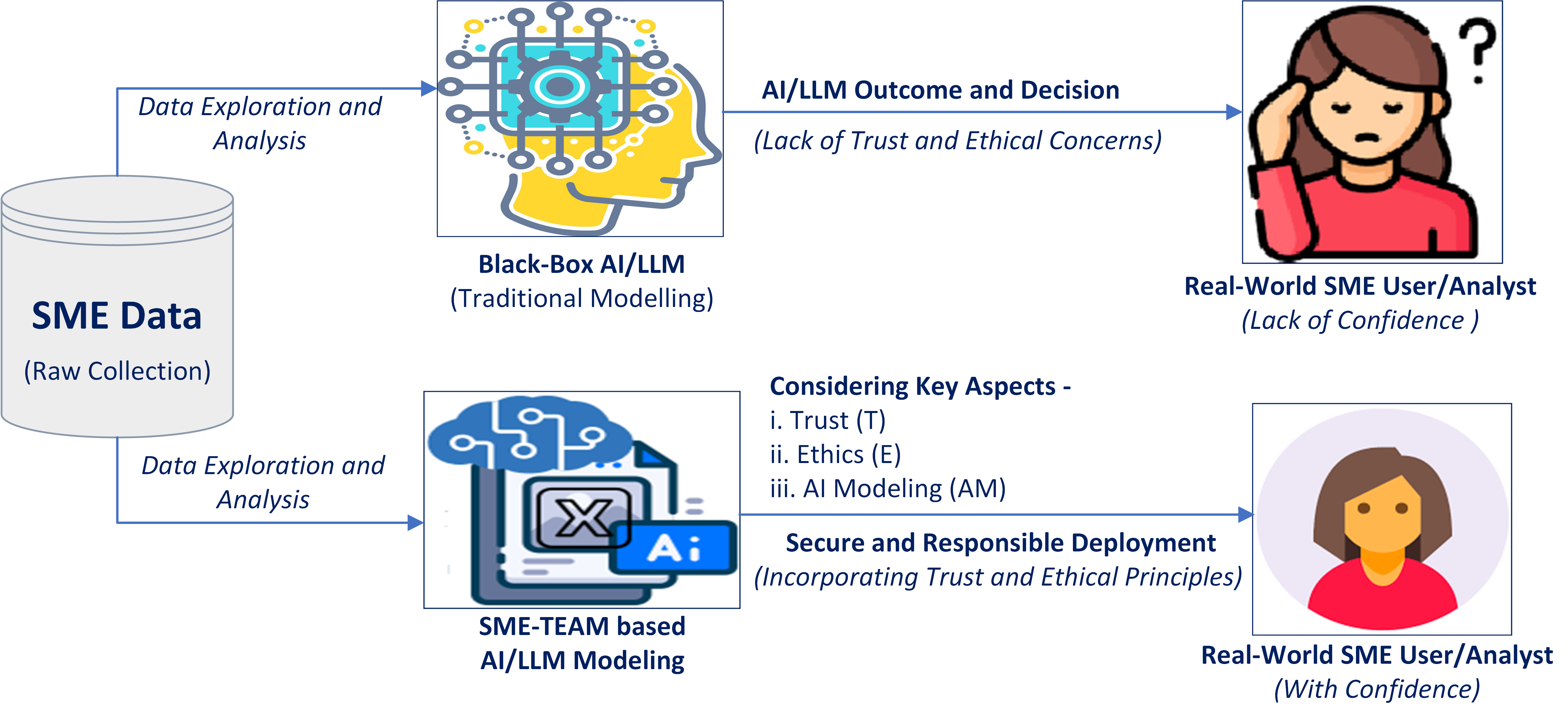}
	\caption{A motivational scenario demonstrating user confidence from the perspective of real-world applicability of SME-TEAM framework, comparing with traditional modeling.}
	\label{fig:SME-TEAM-scenario}
\end{figure} 

Overcoming these challenges requires SMEs to embed trust and ethical principles, discussed briefly in Section \ref{Conceptual Foundation for the Framework}, at the core of their AI adoption strategies. By prioritizing transparency, explainability, accountability, and a secure-by-design approach in SMEs, technological innovation aligns with both societal values and regulatory expectations. Explainable AI mechanisms \cite{sarker2024explainable, lee2025toward, zhao2024explainability} provide insights into decision-making processes that enable stakeholders to build confidence and trust among consumers, employees, and regulators. Human oversight strengthens ethical design principles, which reduce bias, ensure fairness, and align AI applications with moral and societal values. Importantly, sector-specific adaptation, such as bias mitigation in financial services or transparent consent mechanisms in healthcare, ensures that ethical safeguards align with contextual needs. By adopting secure-by-design architectures and maintaining a robust governance strategy, SMEs can balance innovation with responsibility. Ultimately, trust and ethics shift the adoption of these emerging technologies from being a potential risk to becoming a strategic driver of competitiveness, resilience, and sustainable innovation in SMEs.

Building on these insights, this paper contributes to both theory and practice by addressing the often-overlooked intersection of AI adoption, trust, and ethics within the SME context. \textit{First}, it advances theoretical understanding by reframing trust and ethics not as constraints to AI modeling but as strategic facilitators of competitiveness, resilience, and innovation in digitally transforming SMEs, a trade-off illustrated in Figure \ref{fig:trade-off}. \textit{Second}, it introduces a structured, multi-phased framework ``SME-TEAM", Trust (T), Ethics (E) and AI Modeling (AM) for SMEs, with a motivational scenario in Figure \ref{fig:SME-TEAM-scenario}, that bridges high-level ethical principles with actionable practices through its four identified key pillars: Data, Algorithms, Human Oversight, and Model Architecture. \textit{Third}, it extends research implications by demonstrating the framework's potential and applicability across SME sectors, encouraging future studies to examine contextual challenges and strategies for sustainable digital transformation. Collectively, these contributions offer SMEs, policymakers, and tech businesses both scientific and practical guidelines on how to use AI technologies in a responsible, safe, and secure manner.

The remainder of this paper is organised as follows. The significance of the SME-TEAM framework in SMEs is contextualized in Section \ref{Digital Transformation and AI in SMEs: Current Practice}, which explores current practices of digital transformation and AI technologies in SMEs. The theoretical foundations and structural pillars of the proposed SME-TEAM framework are presented in Section \ref{Conceptual Foundation for the Framework}. Section \ref{The Framework: SME-TEAM} outlines the workflow and presents the methodological multi-phased approach. The SME-TEAM framework's practical relevance in a variety of application areas is highlighted in Section \ref{Realizing Potential and Real-World Applicability in SMEs}. Key findings are discussed with research implications in Section \ref{Discussion and Implications for Research}. Finally, the paper is concluded with a summary of theoretical contributions with practical implications in Section \ref{Final Remarks}.

\section{Digital Transformation and AI in SMEs: Current Practice}
\label{Digital Transformation and AI in SMEs: Current Practice}

This section discusses the current practices that are driving the digital transformation of SMEs, the evolving role that AI and LLMs are playing in speeding up this transformation, and the new risk factors and security concerns associated with their trustworthy and responsible adoption.

\subsection{Digital Transformation of SMEs}
The term ``digital transformation" \cite{kohli2025digital, wang2024strengthening, ayaz2025ethical, da2025digital, tripathi2024mapping} typically refers to the strategic integration of digital technologies across all aspects of business operations, thereby transforming the way SMEs operate and provide value.  Its main objective is to tackle persistent problems like SMEs' limited operational efficiency, data fragmentation, lack of skills, dependence on manual processes, and scalability barriers. Through the adoption of digital tools and platforms, SMEs can improve transparency, agility, and data governance while strengthening customer relationship management, streamlining production, and improving supply chain coordination. As a result, digital transformation has become a defining feature of the modern SME landscape - enabling smaller enterprises to modernize operations, enhance efficiency, and compete effectively in digital markets. Increasingly, this transformation is being accelerated by the integration of AI technologies into operational workflows and strategic decision-making, positioning it as a key catalyst for SME innovation, growth, and long-term competitiveness \cite{SMEAustraliaAIAdoption2025}.

Current adoption trends focus on automating repetitive tasks, improving customer engagement with chatbots and recommendation systems, and using data analytics to make well-informed decisions across various business sectors. For instance, Indiani et al. \cite{indiani2025unlocking} investigate how digital transformation unlocks e-commerce potential; Yuwono et al. \cite{yuwono2024information} emphasize the use of digital technologies across various SME sectors like manufacturing, agriculture, and mining; Kohli et al. \cite{kohli2025digital} highlight how it can improve supply chain management; El et al. \cite{el2024leveraging} investigate how it can be used in sustainable agriculture; and Vukovic et al. \cite{vukovic2025ai} and Lu et al. \cite{lu2023developing} concentrate on financial services. Treasure et al. \cite{treasure2019scaffolded} emphasize applications in healthcare, Da et al. \cite{da2025digital} analyze manufacturing competitiveness, De et al. \cite{de2025technology} explore the role of digital twin technology in SMEs, and Graziosi et al. \cite{graziosi2024vision} discuss sustainable additive manufacturing. While these practices demonstrate the potential of digital technologies to enhance efficiency and competitiveness, they also reveal the uneven readiness of SMEs, largely due to gaps in secure, ethical and responsible adoption of AI technologies.

\subsection{Revolution of Today's AI and LLMs}
AI has rapidly progressed from a theoretical concept to a transformative force reshaping industries, business operations, and decision-making. Advances in ML, DL, data analytics, and natural language processing over the last ten years have made it possible for AI systems to process massive datasets, identify complex patterns, and provide remarkably accurate predictive insights \cite{sannabe2022improve, razzaq2025empowering, sarker2024ai}. In addition to increasing productivity, these developments enable businesses to adopt data-driven practices that increase flexibility and competitiveness. Several studies that highlight the crucial significance of AI adoption in enhancing SME performance, innovation, and long-term competitiveness include Wei et al. \cite{wei2022artificial}, Arroyabe et al. \cite{arroyabe2024analyzing}, Carayannis et al. \cite{carayannis2025enhancing}, and Mantri et al. \cite{mantri2023empowering}. LLMs represent a paradigm shift in language-based intelligence within the broader AI ecosystem \cite{zhao2024explainability, naveed2025comprehensive, ferrag2025generative}. Models like GPT and BERT \cite{ferrag2025generative, sarker2024ai}, which are based on sophisticated DL architectures and trained on large text corpora, demonstrate remarkable capabilities in comprehending, producing, and reasoning with natural language. Conversational agents, automated content production, sentiment analysis, summarization, knowledge extraction, and security analytics simply represent some of their uses. For example, Kopka et al. \cite{kopka2024artificial} analyze their wider impact on productivity and innovation in SMEs, Carayannis et al. \cite{carayannis2024empowering} emphasize their potential to improve SME resilience and competitiveness, and Zhang et al. \cite{zhang2025benchmarking} evaluate the role of LLMs in phishing detection for SMEs.

While AI technologies hold significant promise, SMEs often struggle to fully leverage them due to persistent concerns around trust and ethical implications \cite{soomro2025sem}. In the Australian context \cite{SMEAustraliaAIAdoption2025}, adoption rates vary markedly by enterprise size: 82\% among businesses with 200–500 employees, 68\% for those with 20–199 employees, 40\% for firms with 5–19 employees, and only 33\% for micro-enterprises with 0–4 employees. Sectoral differences are also evident, with relatively higher adoption in retail trade (46\%), health and education (45\%), services (43\%), and hospitality (42\%), compared to lower uptake in distribution (31\%), construction (30\%), manufacturing (28\%), and agriculture, forestry, and fishing (19\%). Despite these disparities, AI remains particularly transformative for SMEs, and to fully realise its benefits, it is essential to not only recognise but also proactively address the associated risks.

\begin{table}[h!]
	\centering
	\caption{Key Gaps in AI Adoption for SMEs} 
	\begin{tabular}{p{3.8cm} p{4.5cm} p{6.8cm}}
		\toprule
		\textbf{Gap Category} & \textbf{Description} & \textbf{Implications for SMEs} \\ 
		\midrule
		\textbf{Technical} & Limited explainability, robustness, and generalisation & AI and LLMs often operate as ``black-boxes”, making it challenging to verify results or detect adversarial manipulation. Weak robustness limits deployment in high-risk or regulated environments and erodes trust. \\
		
		\textbf{Ethical} & Insufficient focus on fairness, bias, and privacy & AI systems may reinforce systemic bias, violate privacy standards, or operate without transparency, raising regulatory and social issues. \\
		
		\textbf{Operational} & Poor alignment with workflows and cybersecurity practices & Misalignment between AI tools and business operations can generate inefficiencies, compliance flaws can be revealed, and practitioners or stakeholders may become resistant. \\
		
		\textbf{Human–AI Integration} & Absence of oversight, interpretability, and trust mechanisms & Insufficient human supervision increases the likelihood of errors, security breaches, and poor decision-making by encouraging either distrust of AI results or blind reliance on misleading information. \\
		\bottomrule
	\end{tabular}
	\label{tab:AI-gaps-SMEs}
\end{table}

\subsection{AI Risk Factors and Security Challenges}
To ensure safe and responsible adoption, SMEs need to address the complex array of risk factors and cybersecurity challenges caused by the rapid integration of AI and LLMs into business operations \cite{ode2025social}. AI systems are particularly vulnerable to adversarial attacks, in which carefully manipulated inputs skew outputs and result in incorrect or harmful decisions. The integrity and confidentiality of training datasets can be compromised by threats like data poisoning and model inversion attacks, while LLM-specific risks like hallucinations and prompt injection \cite{ferrag2025prompt} raise the risk of misinformation and unreliable results. In addition to these technical flaws, the use of AI technologies can lead to more general issues like algorithmic bias, opaque decision-making, and an excessive dependence on automation, all of which may undermine stakeholder trust and accountability. SMEs are especially at risk because of their relatively insufficient governance and cybersecurity capabilities. Because of these risk factors and ethical concerns, many SMEs have yet to be ready to adopt AI \cite{cooper2025smes}.

The identified gaps can be broadly classified into four domains: technical gaps, ethical gaps, operational gaps, and human–AI integration gaps, as summarized in Table \ref{tab:AI-gaps-SMEs}. Addressing these gaps is essential to ensure that SMEs can adopt these emerging technologies in a way that is not only innovative and efficient but also safe, secure, and responsible. The significance of incorporating secure-by-design principles, robust data governance, continuous monitoring, and structured human oversight throughout the AI lifecycle is thus emphasized. Our proposed multi-phased framework is specifically designed to bridge these gaps, providing SMEs with actionable guidance across the pre-modeling, in-modeling, and post-modeling stages, discussed briefly in Section \ref{The Framework: SME-TEAM} throughout the AI lifecycle.

\section{Conceptual Foundation for the Framework}
\label{Conceptual Foundation for the Framework}
In this section, we establish the conceptual foundation of our proposed SME-TEAM framework by outlining the core trust and ethical AI principles that underpin responsible adoption. We also map these principles onto the framework's foundational pillars illustrating how ethical values are systematically translated into actionable and context-specific practices for SMEs.

\subsection{Trust and Ethical AI Principles}
A set of core trust and ethical principles that serve as an operational compass and moral foundation needs to guide the adoption of AI technologies in SMEs. Drawing from previous research \cite{duenser2023whom, schmid2023importance, crockett2021building, soudi2024ai}, we outline the following key principles, which are interconnected and mutually supported, to ensure that the adoption of these emerging technologies promotes safe, secure and sustainable outcomes. 

These include: (i) \textit{Fairness and non-discrimination}, which ensures that AI systems do not discriminate and treat all groups equally; (ii) \textit{Accountability and responsibility}, which require clear mechanisms of oversight, liability, and redress for AI-driven outcomes; (iii) \textit{Transparency and explainability}, which make decision pathways interpretable for both technical and non-technical stakeholders; (iv) \textit{Privacy and data protection}, safeguard personal information and ensure compliance with regulatory standards; (v) \textit{Safety and security}, embedding resilience against adversarial attacks, misuse, and system vulnerabilities; (vi) \textit{Human-centredness and oversight}, which emphasize the necessity of retaining meaningful human control, especially in high-stakes contexts; (vii) \textit{Inclusivity and accessibility}, ensure that AI serves diverse users and helps bridge digital divides; (viii) \textit{Trustworthiness and reliability}, which emphasize the importance of delivering consistent, dependable results that inspire user confidence; (ix) \textit{Contestability}, ensures that people, groups, or communities affected by AI systems have prompt opportunities to challenge results or decisions; and last but not least, (x) \textit{Human, societal, and environmental wellbeing}, upholds the idea that the adoption of AI technologies should promote not only business efficiency but also the larger interests of society and the environment. 

Together, these principles offer SMEs concrete recommendations on how to responsibly integrate AI and LLMs, establishing a balance between innovation and responsibility and making sure that digital transformation supports long-term resilience and sustainable growth.

\subsection{Framework Pillars: Mapping the Foundational Principles}
\begin{figure}[ht]
	\centering
	\includegraphics[width=.8\linewidth]{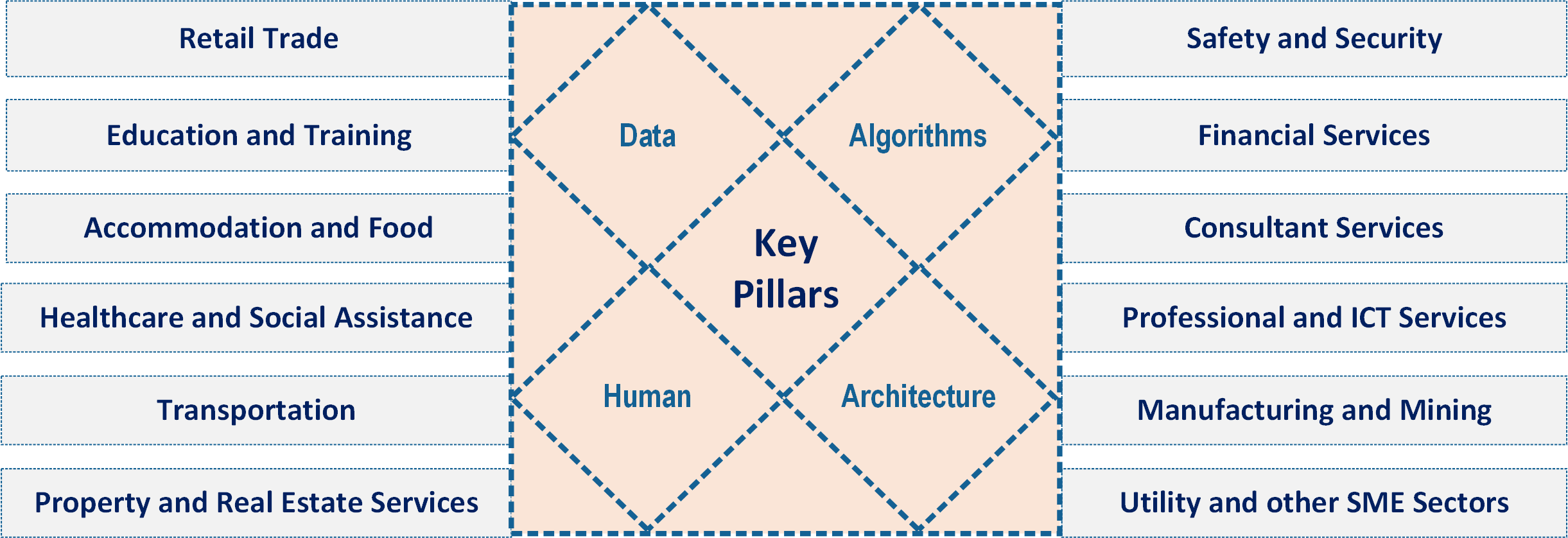}
	\caption{An illustration of SME-TEAM pillars with the example of different SME sectors.}
	\label{fig:SME-TEAM-pillars}
\end{figure}

Based on the theoretical ethical principles outlined above, we have identified four interdependent pillars - (i) Data, (ii) Algorithms, (iii) Human, and (iv) Model Architecture, which collectively translate these key principles of trust and ethics into actionable insights throughout the AI lifecycle in various SME contexts, such as retail, finance, professional services, cybersecurity and other SME sectors, as shown in Figure \ref{fig:SME-TEAM-pillars}. These pillars serve as the structural foundation for translating abstract values into actionable practices, enabling SMEs to effectively utilize the potential of AI technologies, as discussed below.

(i) \textit{Data -} is the foundational pillar of adopting AI technologies in business, as the reliability of AI outputs depends directly on the quality, authenticity, and provenance of data inputs \cite{bettoni2021ai, ha2025empowering}. As SMEs heavily depend on diverse data sources such as internal data, e.g., sales records, customer feedback, supply chains; external data, e.g., industry reports, market trends, social media, economic indicators; as well as real-time feeds, e.g., IoT sensors, sentiment analysis, web analytics \cite{carayannis2025enhancing}, ensuring data integrity through bias detection, anonymisation, and compliance with privacy regulations is crucial. By embedding mechanisms for provenance tracking and validation, SMEs can safeguard against vulnerabilities such as data poisoning, hidden skewness, and regulatory breaches, thereby building confidence in downstream decision-making tasks.

(ii) \textit{Algorithms -} Data is transformed into actionable intelligence by algorithms, which act as the computational engines that drive well-informed decision-making throughout the business operations. To uphold integrity and trust, algorithm design must prioritize robustness, resilience against adversarial manipulation, and fairness. This includes selecting or developing algorithmic models such as ML, DL, data and knowledge mining, and language modeling approaches \cite{sarker2024ai, jafarzadeh2024supporting} that strike a balance between accountability and efficiency for SMEs. To avoid unconscious bias and enhance stakeholder confidence, algorithmic transparency and explainability needs to be established. Furthermore, by incorporating ethical parameters into algorithmic workflows, SMEs can identify and address discriminatory outcomes early, establishing algorithmic accountability as a fundamental component of trustworthy and responsible AI.

(iii) \textit{Human Oversight -} serves as the ethical anchor of the framework, which ensures that moral responsibility and accountability are maintained throughout the AI lifecycle. Even though AI makes automation and scalability potential, SMEs shouldn't trust opaque or completely autonomous systems with important decision-making authority. In high-stakes decision-making, incorporating human-in-the-loop and human-on-the-loop mechanisms maintains domain expertise, ethical reflection, and contextual understanding. This pillar emphasizes that accountability should be transparent and shared, with well-defined escalation pathways that enable human judgment to supersede automated recommendations when needed. This strengthens ethical alignment, safety, and trust in AI-driven operations. 

(iv) \textit{Model Architecture -} provides the technical scaffolding that binds the other pillars discussed above together. Applying secure-by-design principles ensure resilience against cyber threats and adversarial attacks, while explainability \cite{sarker2024explainable, lee2025toward, zhao2024explainability} mechanisms enable transparent validation of model outputs by both technical and non-technical stakeholders. Context-aware architectures further ensure that model outputs are aligned with sectoral requirements and SME-specific goals such as retail, finance, professional services, cybersecurity or other SME sectors. By incorporating these safeguards into the design process, SMEs can achieve a balanced synergy between operational efficiency, security, and ethical values, building user trust in AI-driven tasks and decision-making.

\begin{table}[h!]
	\centering
	\caption{Framework Pillars: Mapping Foundational Principles for SMEs}
	\begin{tabular}{p{3cm} p{5cm} p{8cm}}
		\toprule
		\textbf{Key Pillar} & \textbf{Focus} & \textbf{Implications for SMEs} \\
		\midrule
		\textbf{Data} & Integrity, authenticity, provenance, and compliance & Ensures data integrity through bias detection, anonymisation, and compliance with privacy regulations. Provenance tracking safeguards against poisoning, skewness, and regulatory breaches, building trust in downstream decision-making. \\
		
		\textbf{Algorithms} & Fairness, robustness, accountability, and resilience & Serve as computational engines that transform data into actionable intelligence. Embedding ethical parameters and explainability mechanisms in algorithm design helps SMEs avoid bias, prevent discriminatory outcomes, and enhance stakeholder confidence. \\
		
		\textbf{Human Oversight} & Ethical anchoring, contextual awareness, shared accountability & Human-in-the-loop, on-the-loop systems ensure domain expertise as well as ethical reflection in high-stakes decisions, with accountability shared between humans and AI systems. \\
		
		\textbf{Model Architecture} & Secure-by-design, transparency, alignment with context & Provides technical scaffolding. Secure-by-design principles enhance resilience, transparency and explainability supports stakeholders interpretation, and context-aware architectures align outputs with SME-specific goals and requirements. \\
		\bottomrule
	\end{tabular}
	\label{tab:SME-TEAM-pillars-mapping}
\end{table}

Taken together, these four pillars discussed above map the foundational principles of trust, ethics, and secure AI modeling into actionable practices, as summarized in Table \ref{tab:SME-TEAM-pillars-mapping}. They provide SMEs with a logical framework for integrating accountability, transparency, and resilience across a range of application domains, ensuring that the adoption of AI is not only technically sound but also beneficial to society and sustainable.

\section{The Framework: SME-TEAM}
\label{The Framework: SME-TEAM}
This section introduces the SME-TEAM framework, along with its general structure and essential elements that facilitate the adoption of trust and ethical AI in SMEs. The evaluation guidelines ensure the framework's long-term effectiveness, practical validity, and accountability for a particular use case.

\subsection{Framework Structure and Components}
\begin{figure}[ht]
	\centering
	\includegraphics[width=.8\linewidth, height=12cm]{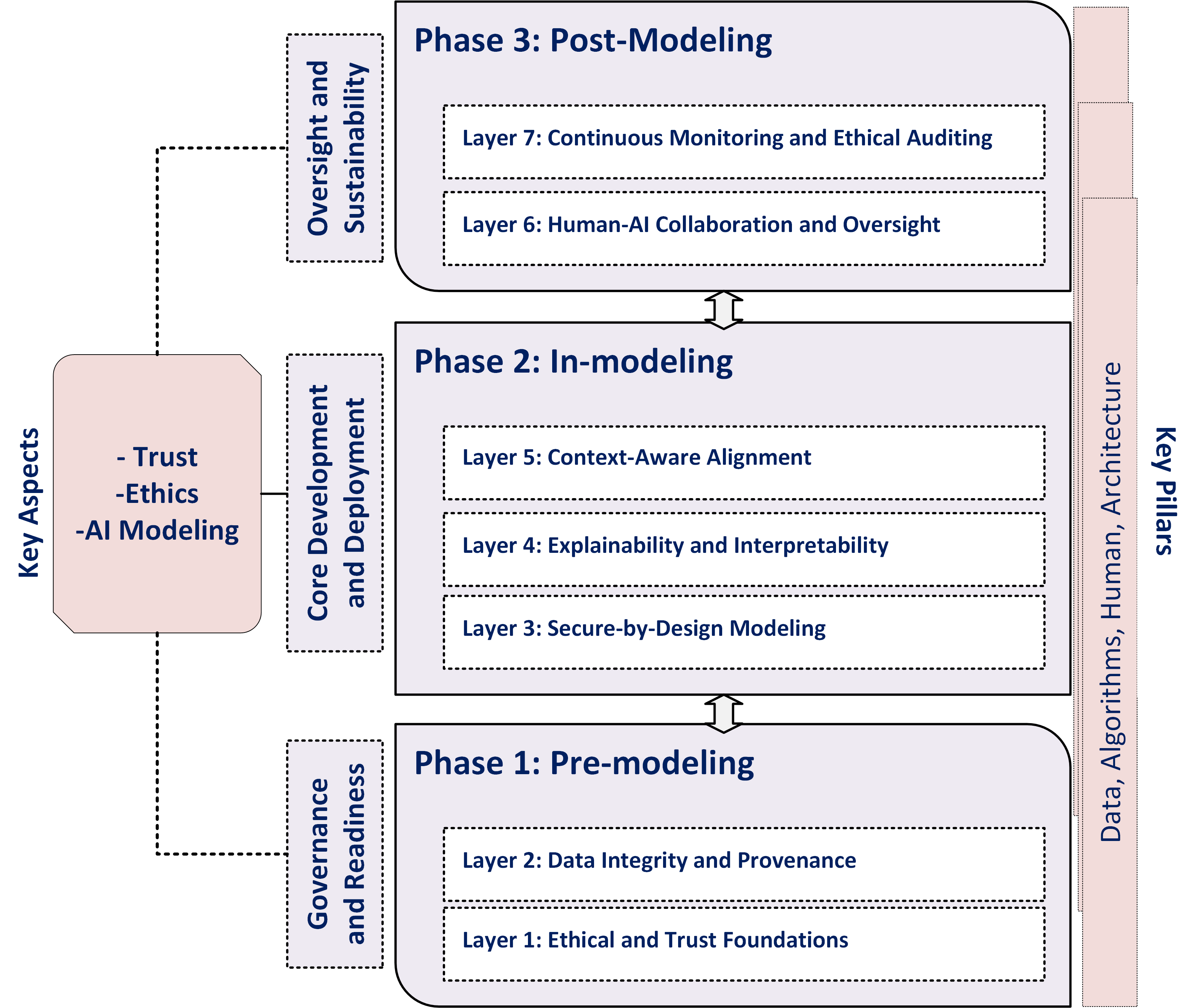}
	\caption{An illustration of SME-TEAM framework.}
	\label{fig:SME-TEAM-framework}
\end{figure}

The SME-TEAM framework is structured into three interconnected phases  as shown in Figure \ref{fig:SME-TEAM-framework}, encompassing seven interdependent layers that collectively embed the principles of trust, ethics, and AI modeling throughout the AI lifecycle. Each phase - pre-modelling, in-modelling, and post-modelling, discussed below - serves a distinct yet complementary function in enabling secure, transparent, and responsible AI adoption within SMEs.

(i) \textit{Pre-Modelling Phase:} This initial pre-modelling phase ``Governance and Readiness" of the framework typically establishes the foundation for secure and responsible use of AI in SMEs by embedding relevant trust and ethical principles before any technical development and integration begins. In a particular business context, for example, financial services, where risks can be disproportionately high, SMEs need to first develop ethical and trust foundations (Layer 1) through lightweight but explicit guidelines that emphasize fairness, inclusivity, accountability, and transparency, ensuring alignment with SME-specific goals and regulatory requirements. At the same time, ensuring data integrity and provenance (Layer 2) becomes essential, with relevant mechanisms such as metadata tagging, anomaly or bias detection, data anonymisation and privacy compliance safeguarding the authenticity and reliability of input datasets. By addressing possible risks and vulnerabilities such as data poisoning, unauthorized manipulation, and hidden skewness early, this pre-modeling phase supports to build a credible baseline of trust and integrity. Collectively, these practices ensure that SMEs approach AI adoption with the necessary ethical, legal, and contextual readiness, avoiding premature data-modeling and experiments, and thus promoting the resilience required for trustworthy and responsible AI systems.

(ii) \textit{In-Modelling Phase:} The in-modelling phase ``Core Development and Deployment" forms the technical kernel of AI adoption within the knowledge value chain \cite{akarslan2024developing}, embedding resilience, transparency, and contextual alignment into model design and deployment. The secure-by-design (Layer 3) practices typically strengthen robustness against security vulnerabilities and threats, anomaly detection failures, and injection attacks, thus safeguarding SMEs from potential risks that could undermine reliability and trust. Simultaneously, model explainability and interpretability (Layer 4) mechanisms address the ``black-box" challenge \cite{sarker2024ai} of AI and LLMs by offering human-centered transparency through decision pathways, counterfactual reasoning, and intuitive visualization that enable both technical and nontechnical stakeholders to verify the model outcome as well as decision-making with confidence. Simultaneously, context-aware alignment (Layer 5) incorporates SME sector-specific requirements to ensure that the outputs minimize unintended consequences while remaining responsible, compliant and relevant across sensitive domains like finance and healthcare. Collectively, these practices in this phase transform AI modeling from a solely technical exercise into a safe, comprehensible, and ethically sound strategy that boosts operational credibility and trust for SMEs.

(iii) \textit{Post-Modelling Phase:} The final post-modelling phase ``Oversight and Sustainability" typically ensures the overall long-term trustworthiness, accountability and adaptability of these emerging and powerful technologies in the SME community. Human-AI collaboration (Layer 6) plays a central role, with decision-making protocols formalizing mechanisms such as human-in-the-loop or human-on-the-loop that balance automation with expert judgement, ensuring clear escalation pathways in high-risk domains such as finance, healthcare, and safety-critical industries. Alongside this, continuous monitoring and ethical auditing (Layer 7) is responsible for adaptive safeguards, including bias re-evaluation, drift detection, compliance checks, and performance tracking, enabling SMEs to adapt dynamically to shifting ethical standards and regulatory requirements in a SME-specific operational environment. All these post-modelling strategies shift deployment from a static endpoint to a proactive, continuous lifecycle, ensuring that AI systems continue to be transparent, responsible, and trustworthy long after they are first implemented.

Together, these seven layers in three different phases discussed above form a well-structured framework to adopt AI technologies in business operations. Thus, the SME-TEAM framework empowers SMEs to utilize AI as a catalyst for innovation, competitiveness, and sustainable growth while managing operational and ethical challenges by integrating trust and ethical principles throughout the entire lifecycle.

\subsection{Evaluation Guidelines}
Both human-centric and automatic evaluation can be employed in the evaluation of the SME-TEAM framework for a specific use case in order to provide a comprehensive evaluation of its operational effectiveness, technical robustness, and ethical validity in SME contexts.

\textit{Human-Centric Evaluation:} This dimension assesses how well the framework promotes ethical compliance, trust, transparency, and usability from the viewpoint of SME stakeholders. It measures important elements like perceived fairness, stakeholder trust, the clarity of model explanations, and the effectiveness of human oversight in AI-assisted decision-making processes through surveys, expert evaluations, and participatory studies. These human-centric observations ensure that the framework will remain to be user-friendly, socially acceptable, and aligned with human values.
	
\textit{Automatic Evaluation:} This approach applies data-driven computational and performance-based metrics to objectively validate the framework's technical and ethical reliability. Model accuracy, precision, recall, and F1-score \cite{sarker2024ai} can be measured, along with feature importance scores, explainability completeness, fairness and bias detection scores, and overall integrity across deployment stages, depending on the nature of the SME-specific use cases.

Together, these two complementary evaluation streams create a robust validation ecosystem that bridges human trust and computational verification. The SME-TEAM framework makes sure that AI adoption in SMEs remains both technically reliable and ethically sound by fusing subjective stakeholder insights with objective performance metrics, defined above.

\section{Realizing Potential and Real-World Applicability in SMEs}
\label{Realizing Potential and Real-World Applicability in SMEs}
This section explores how the SME-TEAM framework drives business economic growth through its potential and real-world applicability, while also promoting innovative and sustainable practices through digital business transformation. Key points include:

\textit{Business Model and Digital Value Creation:}
To generate and capture value, business models in the digital age depend more and more on data-driven strategies and AI-enabled capabilities \cite{jafarzadeh2024supporting, soomro2025sem}. Digital value creation transforms raw data, customer interactions, and operational processes into innovative products, services, and experiences that boost competitiveness \cite{jalil2025influential, carayannis2025enhancing}. By incorporating AI into their business plans, SMEs can increase scalability, create customized products and services, and generate new revenue streams. Digital value creation promotes not only economic growth but also socially conscious and sustainable business practices by integrating trust, transparency, and ethical principles.

\textit{Optimization and Increased Productivity and Services:}
AI can be used to optimize and boost productivity in SMEs by automating repetitive tasks, streamlining operations, and improving decision-making through data-driven insights \cite{jafarzadeh2024supporting, soomro2025sem, kohli2025digital}. When budgets and personnel are limited, businesses could achieve more with fewer resources through predictive analytics, intelligent allocation of resources, and process automation. Security, explainability, and context-aware AI systems can help SMEs reduce costs, retain competitiveness in ever-changing digital markets, and increase productivity while maintaining high-quality service without compromising trust or reliability.

\textit{Enhancing Customer Satisfaction and Engagement:}
SMEs can use AI to improve customer engagement and satisfaction through individualized, trust-based, and responsive interactions. Businesses can anticipate customer needs, customize recommendations, and provide seamless support through chatbots and virtual assistants by utilizing data-driven insights \cite{sarker2024ai}. While ethical protections ensure privacy and fairness, explainability and transparency in AI decisions promote confidence. In addition to increasing customer satisfaction, this ethical approach strengthens engagement, converting potential customers into loyal advocates and offering SMEs a competitive advantage in the digital marketplace.

\textit{Risk management, Cybersecurity and Resilience:}
For SMEs to function safely in an increasingly digital ecosystem, risk management, cybersecurity, and resilience are crucial \cite{ode2025social}. Businesses can detect vulnerabilities early and take proactive measures to combat new threats by integrating secure-by-design architectures, continuous monitoring as well as explainable decision-making. While human oversight ensures accountability in high-stakes decisions, AI-powered analytics \cite{sarker2024ai} improve anomaly detection, fraud prevention, and predictive risk assessment. SMEs can preserve business continuity, build stakeholder trust, and continue to operate securely in a rapidly changing cyber environment by protecting data, upholding regulatory compliance, and strengthening resilience.

\textit{Business Reputation Enhancement:}
Enhancing a company's reputation is essential for SMEs to build credibility, trust, and long-term success in competitive markets. SMEs can demonstrate reliability and trustworthiness by implementing ethical AI practices, maintaining transparency, and protecting customer data. In addition to attracting in customers and potential investors, a solid reputation enhances partnerships and regulatory trust. Providing safe, transparent, and reliable services on a regular basis improves reputation and customer satisfaction. At the end, SMEs could achieve resilience, growth, and a sustainable competitive advantage by leveraging their reputation as a strategic asset.

\textit{Market Expansion and Creating New Employment Opportunities:}
One of the main effects of AI adoption in SMEs is the creation of jobs and market expansion. SMEs can discover new opportunities, customize their products, and enter both domestic and foreign markets by utilizing data-driven insights, predictive analytics, and intelligent decision-making. AI-enabled scalability promotes creative business models, increases competitiveness, and reduces entry barriers. SMEs generate new jobs as they expand \cite{inegbedion2024small, ode2025social, kopka2024artificial}. This growth will be inclusive, sustainable, and in line with long-term societal and economic objectives if ethical and responsible AI practices are integrated.

In summary, it offers SMEs a balanced way to safely use AI while preserving resilience and competitiveness in the rapidly evolving digital economy by integrating productivity gains with customer trust, risk mitigation, and sustainable innovation.

\section{Discussion and Implications for Research} 
\label{Discussion and Implications for Research}

The SME-TEAM framework emphasizes the need for AI technologies that are both technically sound, ethical and contextually practical for SMEs, as they usually have smaller technical teams, limited budgets, and fewer computational resources than large corporations. Many of the current strategies for explainability, bias detection, and adversarial robustness are designed for enterprise-scale environments and necessitate substantial infrastructure and knowledge that SMEs cannot reasonably maintain. Research into resource-efficient, lightweight methods is necessary to address this gap. Examples include robustness mechanisms that can operate on typical SME-level infrastructure, bias detection algorithms customized for SME sector specific datasets, as shown in Figure \ref{fig:SME-TEAM-pillars}, and simplified explainability tools that generate insights that are understandable by SME users and analysts. By adapting innovations to the SME environment, barriers to adoption can be reduced and makes a significant contribution to the economy.

Beyond efficiency, sector-specific interpretability frameworks that transform AI outputs into insights relevant to their application context are essential. Although general-purpose interpretability techniques provide algorithmic transparency, they frequently overlook the operational and ethical realities of particular domains. Interpretability in healthcare needs to be in line with clinical protocols and patient safety standards; in manufacturing, explanations need to emphasize implications for quality assurance and operational efficiency; and in finance, risk models need to be explicable in terms of equity and complying to monetary regulations. Promoting domain-specific interpretability research ensures that AI results are not only transparent but also practical and ethically sound, allowing SME stakeholders to embrace AI-driven decision-making with more assurance.

Another promising direction is the creation of drift detection and adaptive auditing techniques suited to SME environments. Adaptive auditing, in contrast to static compliance checks, facilitates ongoing monitoring, allowing SMEs to remain compliant with evolving legal and ethical requirements without continuously depending on external supervision. By creating lightweight systems that identify ethical risks, data drift, and concept drift in real time, research can progress this and address vulnerabilities before they become systemic failures. In addition to the technical aspect, interdisciplinary cooperation is equally important. Social sciences, business studies, and ethics insights can help reveal how stakeholder values, employee trust, and organizational culture affect the adoption of AI.

While the SME-TEAM framework focuses primarily on conceptualization and methodological underpinnings to ensure a clear articulation of trust, ethics, and AI modeling for SMEs, employing the framework to use in an actual SME setting would further validate its practicality and effectiveness. Future work will include a pilot implementation of this framework for a particular SME use case, e.g., professional service. The empirical analysis and both human-centric and automatic validation as discussed in Section \ref{The Framework: SME-TEAM} will strengthen the evidence base and demonstrate how SME-TEAM supports responsible, trustworthy, and explainable AI adoption in real business contexts.

\section{Concluding Remarks}
\label{Final Remarks}
This section concludes the paper by summarizing both its theoretical contributions, advancing the understanding of trust, ethics, and responsible AI adoption within SME contexts, and its practical implications, which offer actionable guidance for implementing the framework in real-world business environments.

\subsection{Theoretical Contributions}
This paper contributes to the expanding body of research on secure and responsible AI by proposing a structured framework that embeds trust and ethical principles throughout the AI lifecycle in the SME context. The framework advances theory in four key ways. First, it bridges the gap between abstract ethical principles and their operationalisation, illustrating how values such as fairness, transparency, and accountability can be translated into actionable layers of AI adoption. Second, it introduces four conceptual pillars - Data, Algorithms, Human Oversight, and Model Architecture, as theoretical anchors that connect ethical values with practical implementation, offering a robust lens for balancing technical, organisational, and ethical dimensions. Third, it adopts a multi-phased perspective encompassing pre-modelling, in-modelling, and post-modelling stages, thereby providing a holistic understanding of ethical AI deployment across the full pipeline rather than in isolation. Finally, it extends AI governance debates by positioning trust and ethics not as constraints but as enablers of sustainable innovation, resilience, and competitiveness. Collectively, these contributions establish a novel foundation for understanding how SMEs can integrate AI responsibly while ensuring long-term growth and societal value.

\subsection{Practical Implications}
Beyond its theoretical significance, the SME-TEAM framework provides business with a realistic guidelines for adopting AI securely and responsibly. This enables business to leverage automation, predictive insights, and personalisation while maintaining data integrity, privacy standards as well as regulatory compliance. For business leaders, it offers actionable guidance on boosting productivity, opening up new markets, reducing cyber risks, and driving sustainable innovation. Importantly, it places a strong emphasis on the growth of the digital market and the creation of new job opportunities, making sure that transformation promotes workforce development rather than displacement. For policymakers, the framework serves as a scalable blueprint to develop funding plans, digital strategies, and regulatory actions aimed at sector-specific SMEs. Collectively, it illustrates how trust and ethics centred AI adoption can position business and SMEs as resilient, innovative, and competitive actors in the digital economy.

\bibliography{bib-SME}

\section*{Author contributions statement}
Author's contributions: conceptualization and writing - original draft preparation: I.H.S; writing-review and editing: H.J, A.M. and L.M.; project planning and administration: I.H.S and H.J.; All authors reviewed the manuscript.

\section*{Funding}
This work is partially supported by funding from the School of Science, Edith Cowan University, Australia (Project:24862).

\section*{Acknowledgement}
AI tool is used refining the text to improve language precision and readability of the paper, particularly for general readers worldwide.

\section*{Additional information}
Correspondence and requests for relevant materials should be addressed to Iqbal H. Sarker.

\end{document}